\documentclass{amsart}

\usepackage{amsmath,amsthm,amscd,amssymb,eucal,mathrsfs}
\usepackage{amsfonts}
\usepackage{dsfont}
\usepackage{latexsym}
\usepackage{graphicx}
\usepackage{multicol}

\usepackage[all]{xy}

\newtheorem{thm}{Theorem}[section]

\newtheorem{lem}[thm]{Lemma}

\newtheorem{rem}[thm]{Remark}
\newtheorem{dfn}[thm]{Definition}

\newtheorem{alg}[thm]{Algorithm}

\newtheorem*{Satz*}{Satz}

\newcommand{\Alg}{{\rm alg}}
\newcommand{\mathset}[1]{{\left\{#1\right\}}} 
\newcommand{\absolute}[1]{\left\lvert#1\right\rvert}
\newcommand{\norm}[1]{\left\|#1\right\|}

\newcommand{\children}{{\rm  ch}}
\newcommand{\Vertices}{{\rm Vert}}
\newcommand{\Ends}{{\rm Ends}}
\newcommand{\Edges}{{\rm Edges}}

\DeclareMathOperator{\Trace}{Trace}
\DeclareMathOperator{\Intra}{Intra}
\DeclareMathOperator{\Inter}{Inter}
\DeclareMathOperator{\Validity}{Validity}
\DeclareMathOperator{\ValidityIndex}{vi}

\begin{document}

\title{A $p$-adic RanSaC algorithm for stereo vision using Hensel lifting}

\author{Patrick Erik Bradley}
 
\date{\today}

\begin{abstract}
A $p$-adic variation of the Ran(dom) Sa(mple) C(onsensus) method for
solving the relative pose problem in stereo vision is developped.
From two $2$-adically encoded images a random sample of five pairs of corresponding
points is taken, and the equations for the essential matrix are solved 
by lifting solutions modulo $2$ to the $2$-adic integers. A recently devised
$p$-adic hierarchical classification algorithm imitiating the known LBG
quantisation method
 classifies the solutions for all the samples after
having determined the number of clusters using the known intra-inter validity
of clusterings. 
In the successful case, a cluster ranking  will determine the cluster
containg
a $2$-adic approximation to the ``true'' solution of the problem. 
\end{abstract}

\maketitle


\section{Introduction}

High dimensional and sparse encodings of data tend to be ultrametric,
and ultrametric spaces allow certain computational operations, like nearest
neighbour finding, to be carried out very efficiently
\cite{Murtagh-JoC2004}. 
This suggests, for the task of hierarchical classification, an ultrametric
encoding of data from the start. This has the advantage that the hierarchical
structure is uniquely determined by the ultrametric, however at the price of
having to find a suitable encoding in an ultrametric space \cite{Brad-JoC}.
A natural family of ultrametric spaces is given by $\mathds{Q}_p$, the $p$-adic numbers
for any prime number $p$ of choice. A first application of classification
algorithms to $p$-adic data in image segmentation  is described in
\cite{B-PXK2001},
where it was found that the $p$-adic algorithms outperformed their classical
counterparts
in efficiency. In \cite{Brad-pNUAA2009}, it was observed that $p$-adic clustering
algorithms need not change the metric when computing distances between
clusters. 

The task of finding optimal classifications lead to the well-known LBG
algorithm, named after the initials of their authors
\cite{LBG1980}. The clusters from the previous step are split by regrouping
the data around the new ``centres'' in an optimal way.
A direct $p$-adic analogue does not exist. However, if clusters are
interpreted as vertices in the dendrogram for the given data, 
then splitting can be performed by replacing a vertex by its children.
Splitting in direction of highest gain and, after finding the clustering,
determining cluster centres leads to what we call  LBG$_p$ algorithm \cite{Brad-LBGp}.
For LBG, the desired number of clusters has to be pre-specified. For LBG$_p$,
we have a pre-specified upper bound for the cluster number.
This leads to the issue of determining that number or bound. In
\cite{RayTuri1999},
an optimisation scheme for the number of clusters is developped for the $k$-means clustering algorithm.
We propose a $p$-adic adaptation of this 
for choosing optimal clusterings among the LBG$_p$-optimal ones of varying size.

An important algorithm in image analysis is Random Sample Consensus, known as
RANSAC \cite{ransac}. In short, this is a general method for fitting a model to
experimental
data by  randomly sampling  the minimal number of points necessary for 
the fit. Then that set of feasible points is enlarged by adding all nearby
points, i.e.\ those points having a fitted model not much different from the
first fit. This is the consensus set.
The sample with largest consensus set yields the best model prediction.
A variation of this idea would be to perform a classification of
the fitted models for all random samples taken from the data.
Then the biggest cluster (with respect to some measure) contains
in its centre the ``true'' model.
Here, we describe a $p$-adic form of this random sample consensus via
classification using the LBG$_p$ algorithm, applied to the relative pose
problem
in stereo vision as described in the following paragraph. 

The issue of estimating camera motion from two views is classical by now, and
methods from projective and algebraic geometry towards its solution were
employed already at an early stage. The relationship between the views
is established by finding correspondences between point pairs taken from both
images.
The fundamental matrix faithfully encodes  the geometric relationship between
the two images. For normalised cameras, the fundamental matrix coincides with the
so-called essential matrix. In general, the two matrices are related through
the camera calibration. Hence, if the calibration is known, it is sufficient
to estimate the essential matrix in order to  solve the relative pose problem
upto a sign ambiguity.
In order to efficiently use a RANSAC for this task, it makes sense to use 
in each sample the minimal number of point correspondences needed.
This number is known to be five \cite{Kruppa1913}.
Only recently, an efficient solution to the five-point relative pose problem
has been developped \cite{5ptNister}. The first algorithms use eight point
pairs, for which
all equations are linear \cite{8pt}. Seven point pairs require a non-linear
constraint, and so does the method using six points
\cite{7pt-a, 7pt-b,6pt}. The idea for the five point problem
is to first solve the five linear constraints, and then insert the general
solution into the other constraints.
This leads to nine homogeneous cubic equations in four unknowns.
These describe a subspace $S$ of projective three-space $\mathds{P}^3$
over the ground field $K$, 
and the relative pose  problem  has a solution if the  space $S$ is
zero-dimensional.
In that case, it is known to be of cardinality ten
\cite{FM1990}, if multiplicities are
taken into account and all solutions from an algebraic closure of $K$ are allowed.
Nist\'er's approach is to eliminate variables and then numerically solve the
resulting univariate polynomial of degree $10$.
An alternative natural approach is by using Gr\"obner bases
\cite{Stewenius2006}.
A formulation of the problem as a polynomial eigenvalue problem which can be
solved robustly and efficiently
is found in \cite{KBP2008}. 

By considering image coordinates as $2$-adic numbers, e.g.\ through an interval
subdivision process, we can rewrite the equations for the five-point relative
pose problem as polynomials with coefficients in $\mathds{Z}_2$, the $2$-adic
integers.
This allows us to use Hensel's lifting method for their solution, and we
arrive at $2$-adic essential matrices which can be approximated by finite
$2$-adic expansions depending on the desired resolution. In this article,
we take a closer look at the structure of the cubic equations and arrive
at a union of algebraic varieties defined by linear and quadratic equations
which have to be intersected with the remaining cubic constraint over the field $\mathds{F}_2$ in the initial stage
before the lifting. An efficient solution of those particular equations 
with Gr\"obner bases is deferred to 
future work. However, we expect the existing literature on 
Gr\"obner basis methods for equations over finite fields to provide results
which can be ``tuned'' to our situation.
The advantage of using the field $\mathds{F}_2$ is that Gr\"obner basis
computations
become very efficient. Together with our decomposition into quadratic and
linear equations, we expect higher performance in comparison with the 
real methods operating on the undecomposed equations. Using the Hensel lifting
methods
leads us to expect higher robustness. However, these expectations yet await
practical evaluation.

\medskip
The following section collects general facts on $p$-adic numbers which we will
use, and fixes some notation. For the convenience of the reader, we 
prove that $p$-adic vectors can be viewed as $p$-adic numbers in an unramified
extension field of $\mathds{Q}_p$ via an isometric isomorphism.
In Section $3$, we review the LBG$_p$ algorithm and show how one can determine
the number of clusters. Section $4$ explains how to arrive at a $2$-adic
encoding of image coordinates, and then develops the lifting algorithm for the 
five-point relative pose equations. Section $5$ incorporates that algorithm
into RanSaC$_p$, the random sample consensus algorithm via classification.

\bigskip
Some part of this article was presented at the Fourth International Conference on $p$-Adic Mathematical Physics
$p$-ADIC MATHPHYS.2009 which took place near Grodna, Belarus.

\section{Generalities} \label{sec-general}

In this section, we review some facts about $p$-adic numbers for later use. A
general introduction can be found in \cite{Gouvea}, the Haar measure on local fields
 is
introduced
in \cite{Weil1973}.

\subsection{$p$-adic fields}
Let $p$ be a prime number, and
$K$ a  field which is a finite extension field of the field $\mathds{Q}_p$ of
rational $p$-adic numbers. We call
$K$ a {\em $p$-adic field}, and its elements simply {\em $p$-adic numbers}. 
$K$ is a normed field whose norm $\absolute{\ }_K$ extends the $p$-adic norm
$\absolute{\ }_p$ on $\mathds{Q}_p$. 
Let
$\mathcal{O}_K:=\mathset{x\in K\mid \absolute{x}_K\le 1}$ denote the local ring of integers of
$K$. Its maximal ideal $\mathfrak{m}_K=\mathset{x\in K\mid \absolute{x}_K<1}$
 is generated by
a {\em uniformiser} $\pi$. It has the property $v(\pi)=\frac{1}{e}$, 
where $e\in\mathds{N}$ is the ramification degree of $K/\mathds{Q}_p$. 
If $e=1$, then $K$ is called {\em unramified} over $\mathds{Q}_p$.

\smallskip
All elements  $x\in K$ have a $\pi$-adic expansion
\begin{align}
x=\sum\limits_{i\ge-m} \alpha_i\pi^i \label{pi-adic-expansion}
\end{align}
with coefficients $\alpha_i$ in some set $\mathcal{R}\subseteq K$ of
representatives for the residue field
$O_K/\mathfrak{m}_K\cong\mathds{F}_{p^f}$.  
In the case $f=1$, the choice $\mathcal{R}=\mathset{0,1,\dots,p-1}$ 
is quite often made. If $K$ is unramified of degree $n$ over $\mathds{Q}_p$, then $f=n$.

\smallskip
The Haar measure on $K$  will be denoted by $dx$ and is normalised to
$$
\int_{O_K}dx=1.
$$

\subsection{$p$-adic vectors as $p$-adic algebraic numbers}
In this subsection, we show how to consider higher-dimensional $p$-adic data
as one-dimensional data in some appropriate unramified field extension.
As a consequence, any classification method with $p$-adic numbers 
can be applied to $p$-adic vector data.

\bigskip
On the vector space $\mathds{Q}_p^n$  there is the maximum norm
$\norm{\cdot}_{\max}$ given by
$$
\norm{x}_{\max}=\max\mathset{\absolute{x_1}_p,\dots,\absolute{x_n}_p},
$$
where $x=(x_1,\dots,x_n)\in\mathds{Q}_p^n$. The following lemma allows to
consider vectors as one-dimensional objects in some $p$-adic number field:

\begin{lem} \label{1dimvect}
There is an isometric isomorphism between  normed vector spaces:
$$
(\mathds{Q}_p^n,\norm{\cdot}_{\max})\cong (K,\absolute{\cdot}_K),
$$
where $K$ is any unramified extension field of $\mathds{Q}_p$ of degree $n$.
Furthermore, for all $n$ there exists such a $p$-adic field $K$.
\end{lem}

\begin{proof}
{\em Isometry.} Any $x\in K$ has a $p$-adic expansion
\begin{align}
x=\sum\limits_{\nu\ge m}a_\nu p^\nu, \label{pexpansion}
\end{align}
where $p$ retains the uniformising property, because $K$ is unramified over
$\mathds{Q}_p$.
The coefficients $a_\nu\in K$ are taken from a set $\mathcal{R}$ of representatives of
the residue field $\kappa=O_K/ p\,O_K\cong \mathds{F}_{p^n}$, where
$\mathcal{R}$ contains the zero element of $K$.
The residue field, as a vector space, is isomorphic to $\mathds{F}_p^n$.
Hence, the coefficients can be identified with vectors in $\mathds{Q}_p$ whose
entries are taken from the set $\mathset{0,\dots,p-1}$ which represents the
residue field $\mathds{F}_p$. This yields a bijection
between $K$ and $\mathds{Q}_p^n$ which is in fact an isomorphism of vector spaces.
Observe now that the norm
$\absolute{x}_K=p^{-m}$, where $m\in\mathds{Z}$ is the smallest exponent $\nu$ of
$p$
in the $p$-adic expansion (\ref{pexpansion}) of $x$ such that $a_\nu\neq
0$. By the above identification of $x$ with a vector  $(x_1,\dots,x_n)\in\mathds{Q}_p^n$,
this means that 
$m$ is the smallest exponent for which the coefficient is not the zero vector.
Hence,
$$
\absolute{x}_K=p^{-m}=\norm{(x_1,\dots,x_n)}_{\max}
$$
as asserted.

\smallskip\noindent
{\em Existence.} Let $\zeta$ be a primitive $(p^n-1)$-th root of unity. Then
the cyclotomic field $K=\mathds{Q}_p(\zeta)$ is an  unramified extension field
of $\mathds{Q}_p$ of degree $n$ \cite[Prop. 5.4.11]{Gouvea}.  
\end{proof}

Consequently, a set of $n$-dimensional $p$-adic vectors can be treated as data taken from a one-dimensional algebraic
$p$-adic field $K$. Hence, its   dendrogram for the maximum norm 
can be viewed as a subtree of the
Bruhat-Tits tree
for $K$. In particular, the classification methods from \cite{Brad-LBGp,Brad-pNUAA2009} for data from $K$ apply to
this case.

\section{An almost optimal $p$-adic classification algorithm}

After briefly reviewing the $p$-adic variation of the
hierarchical classification algorithm
of \cite{LBG1980}, we show how to determine the optimal number of clusters
in the $p$-adic case. More details on the $p$-adic classification algorithm
can be found in \cite{Brad-LBGp}. The cluster number determination is a
$p$-adic analogue of the method from \cite{RayTuri1999}.



\subsection{LBG$_p$}

In \cite{LBG1980} a clustering algorithm is presented which 
 determines optimal clusterings of given real vector data.
This so-called {\em split-LBG algorithm} constructs cluster centres around
which the data are grouped in an optimal manner.
This method has no direct $p$-adic analogon.
However, \cite{Brad-LBGp} shows an adaptation of the
split-LBG algorithm to $p$-adic data which locally
minimises the expression
$$
E(\mathscr{C},{\bf a})=\sum\limits_{c\in\mathscr{C}}\sum\limits_{x\in C}\absolute{x-a_C}_K,
$$
where $\mathscr{C}$ is a partition {\em (clustering)} of given data $X$, and 
${\bf a}=(a_C)_{C\in \mathscr{C}}$ consists of {\em centres} of clusters $C$. The clusterings are bounded a
priori in size
by $\absolute{\mathscr{C}}\le k$. The method is to
first subdivide given
clusters 
in order to obtain largest decrease in $E(\mathscr{C},{\bf a})$,
and then within the found clustering 
to find   centres in a second step. The centres are characterised by
their minimising property  for $E(\mathscr{C},{\bf a})$ with a given clustering $\mathscr{C}$.
More details can be found in \cite{Brad-LBGp}.
The cluster centres will be used later in Section \ref{ransacp} for obtaining an estimated solution to the 
relative pose problem in stereo vision.

\subsection{Determining the number of clusters}
The LBG$_p$ algorithm as described in the previous subsection uses as input a
pre-specified upper
bound for the number of clusters. Usually, however, this number is not known a
priori.
There are in the literature various methods for finding optimal cluster
numbers.
The application we have in mind is to find a clustering in which one cluster contains the ``best''
approximations
to some unknown quantity, and the other clusters are ``outliers''.
Hence, the ideal clustering should contain compact
clusters which are well separated.
For this reason, we define a $p$-adic version of the intra-inter-validity index
from \cite{RayTuri1999}.

Let $X\subseteq K$ be a finite set, and $\mathfrak{X}_k(X)$ the set of all
verticial clusterings of $X$ with cardinality $\ell\le k$. For a  choice of cluster centres 
${\bf a}=(a_C)_{C\in\mathscr{C}}$ in a given clustering $\mathscr{C}\in\mathfrak{X}_k(X)$, we define
the following quantities:

\begin{align*}
\Intra(\mathscr{C})&=\frac{1}{\absolute{X}}\sum\limits_{C\in\mathscr{C}}\sum\limits_{x\in C}\absolute{x-a_C}_K\\
\Inter(\mathscr{C})&=\min\limits_{C\neq C'\in\mathscr{C}}\absolute{a_C-a_{C'}}_K\\
\Validity(\mathscr{C})&=\frac{\Intra}{\Inter}
\end{align*}

\begin{lem}
$\Intra$ and $\Inter$ do not depend on the choice of cluster centres   ${\bf a}$.
\end{lem}

\begin{proof}
$\Intra$. This follows from the definition of cluster centre, as
$$
\Intra(\mathscr{C})=\frac{1}{\absolute{X}}\ E(\mathscr{C},{\bf a}).
$$

\medskip\noindent
$\Inter$. This follows from the strict triangle inequality, as 
distinct clusters $C\neq C'$ from $\mathscr{C}$
are disjoint.
\end{proof}

The function $\Intra$ measures the compactness of a cluster, whereas $\Inter$
is a measure for the inter-cluster distance. A ``good'' clustering would 
minimise $\Intra$ and maximise $\Inter$. So, an obvious measure combining both
tasks
is given by $\Validity$ as the ratio of both measures.

\begin{dfn}
The quantity
$$
\ValidityIndex_k(X)=\min\limits_{\mathscr{C}\in\mathfrak{X}_k(X)}\Validity(\mathscr{C})
$$
is called the  {\em $k$-th validity index} of the data $X$. 
\end{dfn}

\begin{lem}
If $N=\absolute{X}$, then
$
\ValidityIndex_N(X)=0.
$
\end{lem}

\begin{proof}
This follows from the fact that $\mathfrak{X}_N$ contains the clustering
consisting of $N$ singletons.
\end{proof}

Hence, the ideal cluster number is to be determined by computing the
$k$-th validity index for  $k<<N$. 

\begin{rem} \label{idealclust}
It is to be expected that for genuine data
$X$, the decreasing function $k\mapsto\ValidityIndex_k(X)$
remains constant on some large interval $I$ inside
$\mathset{2,\dots,\absolute{X}}$,
and that the minimum of $\Validity$ on $\mathfrak{X}_k$ with $k\in I$ is attained 
on some clustering $\mathscr{C}\in\mathfrak{X}_k$ such that
$\absolute{\mathscr{C}}<<\absolute{X}$.
\end{rem}

\begin{dfn} \label{idealclustering}
A clustering $\mathscr{C}$ of minimal validity in the sense of Remark
\ref{idealclust}
is called an {\em ideal clustering}.
\end{dfn}

\section{The dyadic $5$-point relative pose equations from stereo vision}

A first $p$-adic formulation of the $n$-point relative pose problem of stereo
vision
is presented in \cite{Brad-5pt2}. There, the equations are derived from a
$p$-adic projective camera model. Here, we first review the $2$-adic image encoding
from a hierarchical
classification point of view, and then propose a refined lifting
algorithm from the reduced equations modulo $2$ to  $\mathds{Z}_2$-rational solutions.

\subsection{$2$-adic image encoding}

Let $R\subseteq\mathds{R}^n$ be the unit hypercube $[0,1]^n$. It can be subdivided by
$n$ hyperplanes parallel to the coordinate hyperplanes into $2^n$ hypercubes 
of equal volume. Each of these smaller hypercubes can be subdivided into even
smaller hypercubes in the same way. Repeating this process yields an infinite
rooted tree $\mathscr{T}$
whose vertices are those  hypercubes, and edges are given by pairs
$(R_n,R_{n+1})$ of hypercubes
where $R_{n+1}$ is one of the parts of $R_n$ obtained in the $n$-th
subdivision step.
The root of $\mathscr{T}$ is given by $R_0=R$, and each vertex has $2^n$ children vertices.

Let $v$ be a vertex of $\mathscr{T}$, and
denote $\children(v)$ the set of its children vertices.
A family $\chi$ of bijections
$$
\chi_v\colon\children(v)\to\mathset{0,1}^n
$$
defines a labelling $\lambda_\chi$ on  $\Edges(\mathscr{T})$, the set of edges of $\mathscr{T}$:
$$
\lambda_\chi\colon \Edges(\mathscr{T})\to\mathset{0,1}^n,\quad (v,w)\mapsto\chi_v(w),
$$ 
and this allows for a $2$-adic encoding of 
$R$, as will be
seen in the following.

First, observe that an end of $\mathscr{T}$, i.e.\ an infinite (injective) path
beginning in $R_0$, corresponds to a decreasing sequence of hypercubes
$$
R_0\supseteq R_1\supseteq R_2\supseteq\dots
$$
having a limit
$$
\bigcap\limits_{\nu\in\mathds{N}}R_\nu=\mathset{r}
$$ 
with a well defined point $r\in R$. This yields an inclusion map
$$
\iota\colon\Ends(\mathscr{T})\to R,
$$
where $\Ends(\mathscr{T})$ is the set of ends of $\mathscr{T}$.

We can view an end $\gamma$ as a tree whose edges $e_0,e_1,e_2,\dots$ are 
directed away from the root $R_0$. They can be numbered by saying that 
$\nu(e)$ is the number of edges on the  path segment $[R_0,o(e)]$,
where $o(e)$ is the origin vertex of $e$.
Now, traversing down a path $\gamma\in\Ends(\mathscr{T})$ and picking up the
 labels on edges $e\in\Edges(\gamma)$ along the path yields a $2$-adic vector,
 as given by the map
$$
\varpi_\chi\colon\Ends(\mathscr{T})\to\mathds{Z}_2^n,\quad \gamma\mapsto\sum\limits_{e\in\Edges(\gamma)}\lambda_\chi(e)\,2^{\nu(e)},
$$ 
and which can be interpreted as an algebraic $p$-adic number in some unramified
$p$-adic field $K$ by Lemma \ref{1dimvect}.

\begin{lem}
The map $\varpi_\chi$ is bijective, and there exists a labelling $\lambda_\chi$
such that
$\iota\circ\varpi_\chi^{-1}$ coincides with the 
map
$$
\mu_2\colon\mathds{Z}_2^n\to R,\quad
a=\sum\limits_{\nu\in\mathds{N}}\alpha_\nu\,2^\nu
\mapsto \sum\limits_{\nu\in\mathds{N}}\alpha_\nu 2^{-(\nu+1)},
$$
with $a_\nu\in\mathset{0,1}^n$.
\end{lem}


\begin{proof}
{\em $\varpi_\chi$ bijective.} A $2$-adic vector in $\mathds{Z}_2^n$ corresponds
uniquely to a sequence $\lambda_0,\lambda_1,\lambda_2,\dots$ of elements from
$\mathset{0,1}^n$. This in turn corresponds uniquely to a path from $R_0$ by
construction of the labels on $\Edges(\mathscr{T})$. Hence, $\varpi_\chi$ is bijective.

\smallskip\noindent
{\em $\mu_2$.} Let $\chi^\mu$ be the  family of bijections
$$
\chi^\mu_v\colon\children(v)\to\mathset{0,1}^n
$$
given by the following construction.
Consider the case $n=1$.
Then $\mathscr{T}$ is a binary tree in which for given vertex $v$
any $w\in\children(v)$ corresponds to the interval $R_w$ which is either the left or the right half of the
interval $R_v\subseteq [0,1]$ corresponding to $v$. 
Now, by defining
$$
\chi^\mu_v\colon\children(v)\to\mathset{0,1},\quad
w\mapsto\begin{cases}0,&\text{$R_w$ is the left half of $R_v$}\\1,&\text{$R_w$ is
  the right half of $R_v$}\end{cases}
$$
we obtain the labelling $\lambda_\mu:=\lambda_{\chi^\mu}$, and the map $\varpi_\mu:=\varpi_{\chi^\mu}$.
We need to prove that 
$$
\bigcap\limits_{v\in\Vertices(\gamma)}R_v=\mathset{\mu_2(\varpi_\mu(\gamma))}
$$
for all $\gamma\in\Ends(\mathscr{T})$. 
Let $\gamma_\nu$ be the segment of $\gamma$ given by the first  edges
$e_0,\dots,e_\nu$.
The terminal vertex of $\gamma_\nu$ corresponds to an interval $R_{\nu+1}$ of length
$\frac{1}{2^{\nu+1}}$.
Let $x_{\nu+1}$ be the left boundary of $R_{\nu+1}$. Inductively, it can be
seen as given by
$$
x_{\nu+1}=x_\nu+\epsilon_\nu \cdot\frac{1}{2^{\nu+1}},
$$ 
where $x_\nu$ is the left boundary of $R_\nu$ and $\epsilon_\nu\in\mathset{0,1}$.
Clearly, it holds true that
$$
\epsilon_\nu=\chi_{v_\nu}^\mu(v_{\nu+1})=\lambda_\mu(e_\nu),
$$
if edge $e_\nu$ is given as $e_\nu=(v_\nu,v_{\nu+1})$.
Hence,
$$
\varpi_\mu(\gamma)=\sum\limits_{\nu\in\mathds{N}}\epsilon_\nu2^\nu,
$$
and it follows that the sequence $(x_\nu)$ converges with respect to the Euclidean
absolute norm to 
$$
\sum\limits_{\nu\in\mathds{N}}\epsilon_\nu2^{-(\nu+1)}=\mu_2(\varpi_\mu(\gamma)),
$$
as asserted.

\smallskip\noindent
The general case follows from the case $n=1$ by applying it to each individual coordinate.
\end{proof}

Since $\mu_2$ is a bijection, it follows that $\iota$ is bijective, too.

\begin{rem}
A rectangular photographic image can be viewed as a rectangular domain in $\mathds{R}^2$.
A digital image, however, will be represented 
for simplicity by an $N\times N$ grid in $\mathds{R}^2$
in which we may assume that $N=2^n$. Hence, the points on the image grid
correspond bijectively to the elements of 
$\mathds{Z}/2^n\mathds{Z}\times\mathds{Z}/2^n\mathds{Z}$,
and the exponent $n$  defines the resolution
of the image. The subdivision process
as before increases each coordinate resolution by $1$,
and only the physical restrictions prevent us from obtaining a $2$-adic image grid
$\mathds{Z}_2\times \mathds{Z}_2$. Hence, we may assume 
two idealised encodings of the digital image square:
the Archimedean one is a real square given by the unit square $[0,1]^2\subseteq\mathds{R}^2$,
and the $p$-adic  encoding is  $\mathds{Z}_2\times\mathds{Z}_2$.
The two ideal encodings are assumed compatible in the sense that the Monna map
$\mu_2\colon\mathds{Z}_2^2\to\mathds{R}^2$ embeds the one into the other.
Physically, the real or $2$-adic coordinates will be approximated in finite resolution 
by a grid isomorphic to $\mathds{Z}/2^n\mathds{Z}$ with varying (and ideally arbitrary) $n$.
\end{rem}

\subsection{The linear and cubic equations}

Assume that there are two views on a static 3D scene taken by cameras with
known calibration. Here, the camera model is projective, and we briefly
explain
how the equations for estimating the geometric relationship between
the two 2D views are derived.
For an introduction to multiple view geometry we refer to \cite{HZ2008}.

A  projective camera is a projective map $\mathds{P}^3\to\mathds{P}^2$.
If two such maps are given, the geometric relationship between the two images
$I$ and $I'$
allow to estimate a recontsruction of the 3D scene. This relationship is given
by
the so-called {\em essential matrix} $E$, a projective $3\times 3$ matrix satisfying

\begin{align}
u_i^TE\,u_i'=0\qquad (i=1,\dots,N), \label{lineq}
\end{align}
where $u_i\in I$ and $u_i'\in I'$ are image points corresponding to the same
point in $3$-space. They are given as vectors with $3$ homogeneous
coordinates.
The equations (\ref{lineq}) are linear in the $9$ unknown entries of $E$.
Reconstruction from $E$ is possible through the factorisation
$$
E=T\cdot R,
$$
where $T$ is a skew-symmetric matrix, and $R$ a rotation.
The matrix $T$ gives the translation in $3$-space from the one camera to the
other,
and $R$ their relative angular orientation. There is an ambiguity given by
the alternative factorisation
$$
E=(-T)\cdot (-R),
$$
but this will not be of concern here.

Since $E$ is a projective matrix, $N=8$ sufficiently general point
correspondences uniquely determine $E$. In fact, there exists a 
reconstruction algorithm which works in this way \cite{8pt}. 
However, that method ignores the fact that an essential matrix is necessarily
of rank $2$. So, a $7$-point algorithm came up replacing one of the equations
in
(\ref{lineq}) by the cubic equation
$$
\det(E)=0
$$
\cite{7pt-a, 7pt-b}.
A $6$-point algorithm is described in \cite{6pt}. It is known that the
minimal number of  point correspondences  necessary for solving
 the relative pose problem is five. We now briefly review
the essentials of the $5$-point algorithm by \cite{5ptNister}.

The first step is to solve the linear system (\ref{lineq}) with $N=5$.
The general solution is of the form
\begin{align}
E=x_1E_1+x_2E_2+x_3E_3+x_4E_4,  \label{lineq-sol}
\end{align}
where $\mathset{E_1,\dots,E_4}$ is a basis for the solution space of (\ref{lineq}).
Here, it is assumed that the $5$ corresponding points are chosen in such a way
that
the rank of the system is five.

In the second step, the matrix (\ref{lineq-sol}) is 
plugged into the non-linear conditions for the essential matrix.
These are given as
\begin{align}
2\cdot EE^TE-\Trace(EE^T)\cdot E&=0  \label{cubeq-trace}
\\
\det(E)&=0 \label{cubeq-det}
\end{align}
This yields $10$ homogeneous cubic equations in the four unknowns
$x,y,z,w$.
The original method by Nist\'er is to set $w=1$, eliminate variables
and obtain a univariate polynomial $f(z)$ of degree $10$.
The zeros of $f(z)$ then lead to maximally ten candidate
essential matrices (after discarding the non-real solutions).

\bigskip
Using the $2$-adic image encoding from the previous subsection, we obtain
for the $5$-point relative pose problem
the same equations (\ref{lineq}), (\ref{cubeq-trace}) and (\ref{cubeq-det}).
The difference is now that the coefficients are $2$-adic expansions of natural
integers, and that the wanted solution is a set of $2$-adic essential matrices. 
In the following subsection, we will explain how this can be obtained 
effectively by Hensel's lifting method.

\subsection{Lifting the  equations to $\mathds{Z}_2$}

The structure of the equations from the previous subsection depends on the 
particular sample of five corresponding pairs of points which in the following
 will simply be referred to as the {\em sample}. The algorithm later on will terminate
either
with a set of solutions or with a resampling routine, meaning that another
set of five corresponding point pairs has to be chosen.

Let $\mathscr{F}\subseteq k[x_1,\dots,x_n]$ be a set of polynomials
with coefficients in a field $k$.
Then the zero set of $\mathscr{F}$ will be denoted as $V(\mathscr{F})$:
$$
V(\mathscr{F}):=\mathset{(\xi_1,\dots,\xi_n)\in k^n\mid
  f(\xi_1,\dots,\xi_n)=0\quad\text{for all $f\in\mathscr{F}$}}.
$$
This is also called  a {\em variety defined over $k$}. 
Let $R$ be a unitary commutative ring contained in the algebraic closure $k^\Alg$ of
$k$, and $V\subseteq k^n$ a variety
defined over $k$.
Then the {\em $R$-rational points} of $V$ are those points of $V_{k^\Alg}$ lying in
$R^n$,
where $V_{k^\Alg}\subseteq (k^{\Alg})^n$ is $V(\mathscr{F})$ seen as a variety
defined over $k^\Alg$.
In particular, we will  speak of
$k$-rational points of a variety defined over $k$. The set of $R$-rational points of $V$ will be denoted by $V(R)$.

\subsubsection{The linear equations} \label{lift-lineq}
Assume that  $N=5$. The simplest case for applying Hensel's lifting method 
is that, after dividing off each equation the least common divisor of the coefficients, 
the linear equations (\ref{lineq}) are linearly independent modulo
$2$.
Then, by the multivariate linear Hensel lemma, e.g.\ \cite[Thm.\ 2]{Brad-5pt2},
there is a unique lift of a basis of the solution space of (\ref{lineq}) 
modulo $2$ to a  $\mathds{Z}_2$-basis of the solution space of
(\ref{lineq}).
The constructive nature of the proof yields a lifting algorithm.

\subsubsection{The cubic equations}

The system (\ref{cubeq-trace}) is
 modulo $2$ of the form
\begin{align}
\Trace(EE^T)\cdot E\equiv 0\mod2 \label{trace-mod2}
\end{align}
By construction, the entries of $E$ are zero or homogeneous linear polynomials
in $x_1,x_2,x_3,x_4$. As in \ref{lift-lineq}, we assume that $E$ is not the zero
matrix modulo $2$.
In any case, the diagonal elements are zero or squares of linear forms. Hence,
the polynomial
$$
Q(x_1,x_2,x_3,x_4)=\Trace(EE^T)\mod 2
$$
is a sum of squares:
$$
Q=\sum\limits_{i=1}^4\alpha_1 x_i^2=L^2,
$$
where $\alpha_1,\dots,\alpha_4\in\mathds{F}_2$ and
$$
L=\sum\limits_{i=1}^4\alpha_i x_i
$$
is linear. From this it follows that  (\ref{trace-mod2})
is of the form
\begin{align}
L^2\cdot L_i=0,\qquad i=1,\dots,9, \label{cublin-mod2}
\end{align}
with $L_i$ either zero or linear modulo $2$.
If we assume  that the $\mathds{F}_2$-basis $E_1,\dots,E_4$ 
of (\ref{lineq}) modulo $2$ is
read off a staircase normal form for the reduced linear system of equations,
we observe that the matrix $E\mod 2$ contains 
 four entries
consisting precisely of the variables $x_1,x_2,x_3,x_4$.
Hence, 
the four equations $L^2\cdot x_i=0$ are among (\ref{cublin-mod2}),
and it follows that the solution of that system is given by
\begin{align}
L=0 \label{reduced-mod2}
\end{align}
over the finite field $\mathds{F}_2$. However, the variety $V(L)$ defined by
(\ref{reduced-mod2})
is a hyperplane in the projective space
$\mathds{P}^3_{\mathds{F}_2}$, whereas (\ref{cubeq-trace}) defines
a curve in $\mathds{P}^3_{\mathds{Q}_2}$
 for a generic sample of five corresponding point pairs.

In any case, by the multivariate Hensel lemma \cite[Thm.\ 1]{Brad-5pt2}
 the points of the set
$$
V=\mathset{x\in V(L)\cap V(\det(E)\!\!\!\!\mod 2)\mid \nabla\det(E)(x)\not\equiv 0\mod
2}
$$
are uniquely liftable to $\mathds{Z}_2$-rational points  in $\mathds{P}^3$,
as long as $\det(E)$ is modulo $2$ not the zero polynomial.

In order to obtain more liftable points, we take a closer look at the variety
given by (\ref{cubeq-trace}). Write to this aim
\begin{align}
E=(E_{ij})=\begin{pmatrix}e_1\\ e_2\\ e_3\end{pmatrix}, \label{rows4E}
\end{align}
where $e_i$ are the rows of $E$. Then (\ref{cubeq-trace}) translates to
\begin{align}
(2EE^T-\Trace(EE^T)\cdot\mathds{1})\cdot E=0, \label{tracecondfact}
\end{align}
where $\mathds{1}$ is the unity $3\times 3$-matrix.
By multiplying from the right with some invertible matrix, we may replace the 
rightmost matrix in (\ref{tracecondfact}) by a  triangular matrix
$$
T=\begin{pmatrix}T_{11}&T_{12}&T_{13}\\0&T_{22}&T_{23}\\0&0&T_{33}\end{pmatrix}.
$$
Using (\ref{rows4E}),
this is equivalent to the system
\begin{align*}
T_{11}\begin{pmatrix}e_1^2-e_2^2-e_3^2\\2e_1e_2\\2e_1e_3\end{pmatrix}=0 
\\
T_{12}\begin{pmatrix}e_1^2-e_2^2-e_3^2\\2e_1e_2\\2e_1e_3\end{pmatrix}
+T_{22}\begin{pmatrix}2e_1e_2\\e_2^2-e_1^2-e_3^2\\2e_2e_3\end{pmatrix}=0 
\\
T_{13}\begin{pmatrix}e_1^2-e_2^2-e_3^2\\2e_1e_2\\2e_1e_3\end{pmatrix}
+T_{23}\begin{pmatrix}2e_1e_2\\e_2^2-e_1^2-e_3^2\\2e_2e_3\end{pmatrix}
+T_{33}\begin{pmatrix}2e_1e_3\\2e_2e_3\\e_3^2-e_1^2-e_2^2\end{pmatrix}=0 
\end{align*}
where we use the sloppy notation $e^2=ee^T$ and $e_ie_j=e_ie_j^T$.
This can be simplified to
\begin{align}
T_{11}\begin{pmatrix}e_1^2-e_2^2-e_3^2\\2e_1e_2\\2e_1e_3\end{pmatrix}=0 \label{tracemat1}
\\
T_{11}T_{22}\begin{pmatrix}2e_1e_2\\e_2^2-e_1^2-e_3^2\\2e_2e_3\end{pmatrix}=0 \label{tracemat2}
\\
T_{11}T_{22}T_{33}\begin{pmatrix}2e_1e_3\\2e_2e_3\\e_3^2-e_1^2-e_2^2\end{pmatrix}=0 \label{tracemat3}
\end{align}
Assume now that in (\ref{tracemat1})-(\ref{tracemat3})
 the least common divisor of the coefficients
for each row in each equation (or just the highest common power of $2$) has been divided off.
The equations 
can now successively be simplified in a straightforward manner.
This leads to
 a union of at most $14$ varieties $W_i$, each of which is defined by quadratic and linear equations.
We call this the {\em sampled five-point trace variety} $W$. 
Of interest is the case that $V:=W\cap V(\det(E))$ has dimension one.

Notice that 
$$
\det(E)=0\Leftrightarrow \det(A)=0,
$$
where $A=(A_{ij})$. 
We set
$$
V_i:=W_i\cap V(\det(A))=V(\mathscr{F}_i)\subseteq\mathds{A}^4,
$$
where $\mathscr{F}_i$ is the set of linear and quadratic polynomials from
above, together with the cubic polynomial $\det(A)$.
The sampled five points lead to finitely many candidate essential matrix
if and only if the image $\bar{V}_i$ under the canonical map
$$
\mathds{A}^4\setminus\mathset{0}\to\mathds{P}^3
$$
is zero-dimensional for all $i$ such that $V_i\setminus\mathset{0}\neq\emptyset$.
The dimension can be checked on each affine chart
$x_i\neq 0$. For this, we denote by
$$
\mathscr{F}_i^j:=\mathset{f|_{x_j=1}\mid f\in\mathscr{F}_i}
$$
and
$$
V_i^j:=V(\mathscr{F}_i^j)\subseteq \mathds{A}^3
$$
the affine piece of $V_i$ given by $x_j=1$.

\medskip
If $\mathscr{F}$ is a set of polynomials with coefficients in $\mathds{Z}_2$, 
then $\mathscr{F}\mod 2$ means the set consisting of the
polynomials from $\mathscr{F}$ with coefficients modulo $2$.
By $V(\mathscr{F}) \mod 2$ we mean the zero set of $\mathscr{F}\mod 2$
defined over $\mathds{F}_2$. We declare $$\dim\emptyset:=-1,$$
and  arrive at the following lifting algorithm:

\begin{alg} \label{5ptlift}
{\em Input.} The reduced matrix $A$ with coefficients in $\mathds{Z}_2[x_1,x_2,x_3,x_4]$.

\smallskip\noindent
{\em Step $1$.} Compute for all $i$ the set  $\mathscr{F}_i$ as above, and 
$\mathscr{F}_i\mod 2$. 
If for some $i$ the latter contains a non-zero polynomial, then
  continue. Otherwise, resample.

\smallskip\noindent
{\em Step $2$.} Compute for all $i$ 
the dimension $d_i:=\dim(\bar{V}_i\mod 2)$
on each affine piece $V_i^j\mod 2$, $j=1,\dots,4$.
If all $d_i\le 0$, then continue, otherwise resample.

\smallskip\noindent
{\em Step $3$.} 
 Compute all $\mathds{F}_2$-rational points of
$V_i\mod 2$, and
for all such $\omega\in V_i\mod 2$ the quantity
$\nabla f(\omega)$, where $f\in\mathscr{F}_i\mod 2$.
If some value is 
$$
\nabla f(\omega)\not\equiv 0 \mod 2,
$$
then lift for that $f$ and collect all lifts in the set $\tilde{V}_i$.
If $\tilde{V}_i\neq\emptyset$, then continue. Otherwise, resample.

\smallskip\noindent
{\em Step $5$.} Test for all $i$ and all $v\in\tilde{V}_i$
whether all quantities
$f(v)$ with $f\in\mathscr{F}_i$ are zero. Collect all positively
tested $v$ in $S\subseteq\mathds{A}^4$.  

\smallskip\noindent
{\em Output.} The  lifted finite  set 
$$\bar{S}\subseteq\mathds{P}^3(\mathds{Z}_2)$$
of $\mathds{Z}_2$-rational  solutions.
\end{alg}

Observe that there is a dimension computation in Step $2$, and a solution set
computation in Step $3$, both for equations modulo $2$. These can be effected with Gr\"obner basis methods,
as described e.g.\ in \cite[Cor.\ 3.7.26]{CoCoA1}.

\section{$p$-adic random sample consensus via classification} \label{ransacp}

In this section, we incorporate Algorithm \ref{5ptlift} into a sampling
scheme in which random samples of five point-pairs are taken, and the output
of the lifting algorithm is collected, and then classified. The idea is that
in the end, a pronounced cluster will appear in the classification which
contains among its central elements the ``true'' essential matrix.
Our approach differs from the original RANSAC \cite{ransac} in that we fix the
number of samplings instead of the cardinality of the consensus set,
and that we perform
a hierarchical classification of the solutions from each sample. The consensus
set
corresponds here to one of the  clusters in the classification.  

\bigskip
Let $K$ be a $p$-adic field.  

\begin{dfn}
Let $C\subseteq K$ be a cluster, and let $A\subseteq C$
be the subset of of all central elements with respect to $E$.
Then the rooted tree $\mathfrak{S}(C):=T^\dagger(A\cup\mathset{\infty)}$ is called the {\em central spine} of
$C$. 
\end{dfn}

Since all central elements branch off the tips of
$\mathfrak{S}(C)$,
the central spine of a cluster $C$ says something about the distribution of
the data within $C$. 

\medskip
We define the {\em densitiy} of a verticial cluster $C$  as
$$
\delta(C):=\begin{cases}\displaystyle
\frac{\absolute{C}-1}{\mu(C)},&\absolute{C}>1\\0,&\text{otherwise}
\end{cases}
$$
where 
$$
\mu(C)=\int\limits_{K}1_{\mathds{D}_C}\,dx,
$$
with $\mathds{D}_C\subseteq K$ being the smallest disk containing $C$
and 
$$
1_{\mathds{D}_C}\colon K\to\mathds{R},\quad x\mapsto\begin{cases}1,&x\in C\\0,&\text{otherwise}\end{cases}
$$
the characteristic function.

\bigskip
The following algorithm is a $p$-adic analogon of a variation of the 
Ran(dom) Sa(mple) C(onsensus) algorithm \cite{ransac}
applied to the problem of estimating the essential matrix from two images. In this variation,
the consensus is established by a hierarchical classification of the
collected solutions for the equations given by the sampled five-tuples
of corresponding image points.
In order to establish the ``winning'' cluster, we consider each sample as
casting upto ten votes. Then we establish a ranking of clusters according to
the following criteria:

\bigskip
\begin{center}
\begin{enumerate}
\item majority of votes
\item highest density
\item highest precision
\end{enumerate}
\end{center}

\bigskip
These criteria are to be taken in that order, i.e.\ the clusters are ranked
according to criterion (1). Ties are first broken using criterion (2),
and then with criterion (3). This means a ranking of clusters

\bigskip
\begin{center}
\begin{enumerate}
\item according to their size
\item according to $\delta(C)$
\item according to $\mu(C_c)$,
\end{enumerate}
\end{center}

\bigskip\noindent
where the $C_c$ is given by the following definition:

\begin{dfn}
Let $C\subseteq K$. Then the {\em central cluster} in $C$ is the smallest
verticial subcluster of $C$ containing the centres of $C$.
\end{dfn}

\begin{rem}
In the case that the differences in size, density or precision are small
among the high ranked clusters, it makes sense to allow almost equally
ranked clusters to be considered as ties and to use the next criterion to
break them.
\end{rem}

Fix an isometric isomorphism
$(K,\absolute{\cdot}_K)\cong(\mathds{Q}_2^{3\times 3},\norm{\cdot}_{\max})$,
where $K$ is an unramified extension field of $\mathds{Q}_2$,
and $\norm{\cdot}_{\max}$ is the maximum of the $2$-adic norms of all matrix entries.

\begin{alg}[RanSaC$_p$]
{\em Input.} Numbers $k,m,n,N\in \mathds{N}$  and a finite set $X\subseteq I\times I'$ of pairs of corresponding
points
in two images $I$ and $I'$ represented as $2$-adic vectors.

\medskip\noindent
{\em Step 1.} Sample five random elements of $X$.
If the equations (\ref{lineq}) modulo $2$ have rank $5$, then solve these by
lifting a basis modulo $2$
to a basis modulo $2^n$. If $n>>0$, then the lift yields  exact solutions in
$\mathds{Z}_2$.
If modulo $2$ the rank is smaller than five, then resample.

\medskip\noindent
{\em Step 2.} Perform Algorithm \ref{5ptlift} by lifting to solutions modulo
$\mathds{Z}/2^m\mathds{Z}$, where $m$ is the desired precision.
 Obtain a set of  approximate candidate essential matrices
with entries in $\mathds{Z}/2^m\mathds{Z}$. If that set is non-empty, then
continue.
Otherwise, resample.

\medskip\noindent
{\em Step 3.} Repeat Steps $1$ and $2$ successively $N$ times, and obtain an accumulated set $\mathcal{E}$
of approximate candidate essential matrices from each repetition.

\medskip\noindent
{\em Step 4.}
Use the LBG$_p$ algorithm over $K$ for obtaining $\le \ell$ clusterings of $\mathcal{E}$
with $\ell =2,\dots,k$. Determine the ideal clustering(s), in the sense of
Definition \ref{idealclustering}, within $\mathfrak{X}_k$.
In these, determine the winning  clusters,
 their centres and central spines.

\medskip\noindent
{\em Output.}
A set of approximate central essential matrices.
\end{alg}

\begin{rem}
The desired result of an instance of RanSaC$_p$ applied to genuine image data
would be a clustering in which there is one pronounced cluster $C$ having a central spine which is a  path segment of length $n>>0$. In this case, the central elements of $C$ would 
yield one single candidate essential matrix $E\in(\mathds{Z}/2^n\mathds{Z})^{3\times 3}$
approximating the ``true'' $2$-adic essential matrix for the particular stereo image problem,
while the other clusters could be considered as ``outliers''.
In general, noise in the image will lead to less pronounced clusters and central spines with branching.
The former
can lead to wrong estimates for $E$, and the latter means that the approximated essential matrix can be less precise than in an ideal setting.
\end{rem}

\section{Conclusion}

The $p$-adic classification algorithm LBG$_p$ is incorporated into a random
sample consensus algorithm via classification (RanSaC$_p$) in order to 
efficiently solve the five-point relative pose problem in stereo vision.
The equations occurring in the relative pose problem are 
derived from a $2$-adic encoding of image coordinates, decomposed and then solved
with Hensel's lifting method. The cluster number is determined with a $p$-adic
version of an intra-inter-validity measure originally developped for
$k$-means.
The proposed solution for the essential matrix lies in the centre of the most
significant
cluster.

\section*{Acknowledgements}

The author would like to thank the organisers of the wonderful 
Fourth International Conference on $p$-Adic Mathematical Physics
$p$-ADIC MATHPHYS.2009 which took place near Grodna, Belarus. Thanks to Sven
Wursthorn for a brief introduction to \cite{ransac} applied to stereo vision.


\begin{thebibliography}{xyz}
\bibitem{B-PXK2001}J. Benois-Pineau, A.Yu. Khrennikov, N.V. Kotovich.
{\em Segmentation of Images in $p$-Adic and Euclidean
  Metrics}. Dokl. Math., 64, 450--455 (2001)


\bibitem{Brad-5pt2}P.E. Bradley. {\em A dyadic solution of relative pose
  problems}. Preprint. arXiv:0908.1919v3 

\bibitem{Brad-LBGp}P.E. Bradley. {\em On $p$-adic classification}.  $p$-Adic
  Numbers, Ultrametric Analysis, and Applications. In press.

\bibitem{Brad-JoC}P.E Bradley. {\em Degenerating Families of Dendrograms.}
J.  Classif., 25, 27--42 (2008)

\bibitem{Brad-pNUAA2009}P.E. Bradley. {\em Mumford dendrograms and discrete
  $p$-adic symmetries}. $p$-Adic Numbers, Ultrametric Analysis, and Applications, Vol. 1, No. 2 (2009), 118--127. 

\bibitem{FM1990}O.D. Faugeras and S. Maybank. {\em Motion from point matches:
  multiplicity of solutions.} IJCV, 4, 225--246 (1990)

\bibitem{ransac}M.A. Fischler and R.C. Bolles. {\em Random Sample Consensuns:
  A paradigm for model fitting with applications to image analysis and
  automated cartography.}  Comm. of the ACM, 24, 381--395 (1981)

\bibitem{Gouvea}F.Q. Gouv\^{e}a. {\em $p$-adic Numbers. An Introduction}. 2nd
  ed. Springer (2003)

\bibitem{7pt-a}R. Hartley. {\em Projective reconstruction and invariants from
  multiple images.} T-PAMI. 16, 1036--1040 (1994)

\bibitem{HZ2008}R. Hartley and A. Zisserman. {\em Multiple view geometry in
  computer vision.} 2nd ed. Cambridge University Press (2008)

\bibitem{7pt-b}T.S. Huang and A.N. Netravali. {\em Motion and structure from
  feature correspondence: a review.} Proc. IEEE, 82, 252--268 (1994)


\bibitem{CoCoA1}M. Kreuzer and L. Robbiano. {\em Computational Commutative
  Algebra 1}. Springer (2000)

\bibitem{Kruppa1913}E. Kruppa. {\em Zur Ermittlung eines Objektes aus zwei
  Perspektiven mit innerer Orientierung.} Sitz.-Ber. Akad. Wiss., Wien,
  math. naturw. Kl. Abt. IIa, 122, 1939--1948 (1913)

\bibitem{KBP2008}Z. Kukelova, M. Bujnak, T. Pajdla {\em Polynomial eigenvalue solutions to the $5$-pt and $6$-pt relative pose problems}. BMVC 2008, Leeds, UK, September1-4 (2008)

\bibitem{LBG1980}Y. Linde, A. Buzo, R.M. Gray. {\em An
  Algorithm for Vector Quantizer Design.} IEEE Trans. Commun., 28, 84--95 (1980)

\bibitem{8pt}H.C. Longuet-Higgins. {\em A computer algorithm for
  reconstructing a scene from two projections.} Nature, 293, 133--135 (1981)


\bibitem{Murtagh-JoC2004}F. Murtagh. {\em On ultrametricity, data coding,
  and computation.} Journal of Classification, 21, 167--184 (2004) 


\bibitem{5ptNister}D. Nist\'er. {\em An efficient solution to the five-point
  relative pose problem.} IEEE T-PAMI, 26, 167--184 (2004)


\bibitem{6pt}J. Philip. {\em A non-iterative algorithm for determining all
  essential matrices corresponding to five point pairs.} Photogrammetric
  Record, 15, 589--599 (1996)

\bibitem{RayTuri1999}S. Ray and R.H. Turi. {\em Determination of number of
  clusters in $K$-means clustering and application in colour image
  segmentation.}
In: Proc. ICAPRDT'99, Calcutta, India (1999)

\bibitem{Stewenius2006}H. Stew\'enius, C. Engels, D. Nist\'er. {\em Recent
  developments on direct relative orientation.} ISPRS J. Photogrammetry and
  Remote Sensing, 60, 284--294 (2006)

 
\bibitem{Weil1973}A. Weil. {\em Basic number theory.} 2nd ed. Springer (1973) 
\end{thebibliography}
\end{document}